\documentclass[runningheads]{llncs}
\usepackage{graphicx}
\usepackage{comment}
\usepackage{amsmath,amssymb} 
\usepackage{color}

\usepackage{bm}
\usepackage{siunitx}
\usepackage{caption}
\usepackage{cite}
\usepackage{ctable}
\usepackage{hhline}
\usepackage{booktabs}
\usepackage{csquotes}
\usepackage{pifont}
\usepackage{multirow}
\usepackage{booktabs}
\newcolumntype{L}[1]{>{\raggedright\let\newline\\\arraybackslash\hspace{0pt}}m{#1}}
\newcolumntype{C}[1]{>{\centering\let\newline\\\arraybackslash\hspace{0pt}}m{#1}}
\newcolumntype{R}[1]{>{\raggedleft\let\newline\\\arraybackslash\hspace{0pt}}m{#1}}
\usepackage{hyperref}

\makeatletter
\newcommand*{\rom}[1]{\expandafter\romannumeral #1}
\def\blfootnote{\xdef\@thefnmark{}\@footnotetext}
\makeatother

\begin{document}
\pagestyle{headings}
\mainmatter

\title{Pose2Mesh: Graph Convolutional Network \\ for 3D Human Pose and Mesh Recovery \\ from a 2D Human Pose}

\titlerunning{Pose2Mesh}
%
\author{Hongsuk Choi$^{*}$\and
Gyeongsik Moon$^{*}$\and
Kyoung Mu Lee}

\authorrunning{H. Choi et al.}
%
\institute{ECE \& ASRI, Seoul National University, Korea\\
\email{\{redarknight,mks0601,kyoungmu\}@snu.ac.kr}}
\maketitle

\begin{abstract}
Most of the recent deep learning-based 3D human pose and mesh estimation methods regress the pose and shape parameters of human mesh models, such as SMPL and MANO, from an input image.
The first weakness of these methods is an appearance domain gap problem, due to different image appearance between train data from controlled environments, such as a laboratory, and test data from in-the-wild environments.
The second weakness is that the estimation of the pose parameters is quite challenging owing to the representation issues of 3D rotations.
To overcome the above weaknesses, we propose Pose2Mesh, a novel graph convolutional neural network (GraphCNN)-based system that estimates the 3D coordinates of {\em human mesh vertices} directly from the {\em 2D human pose}.
The 2D human pose as input provides essential human body articulation information, while having a relatively homogeneous geometric property between the two domains. 
Also, the proposed system avoids the representation issues, while fully exploiting the mesh topology using a GraphCNN in a coarse-to-fine manner.
We show that our Pose2Mesh outperforms the previous 3D human pose and mesh estimation methods on various benchmark datasets.\blfootnote{* equal contribution}
The codes are publicly available \footnote{\url{https://github.com/hongsukchoi/Pose2Mesh_RELEASE}}.

\end{abstract}

\section{Introduction}
3D human pose and mesh estimation aims to recover 3D human joint and mesh vertex locations simultaneously.
It is a challenging task due to the depth and scale ambiguity, and the complex human body and hand articulation. 
There have been diverse approaches to address this problem, and recently, deep learning-based methods have shown noticeable performance improvement.

Most of the deep learning-based methods rely on human mesh models, such as SMPL~\cite{loper2015smpl} and MANO~\cite{romero2017mano}.
They can be generally categorized into a model-based approach and a model-free approach.
The model-based approach trains a network to predict the model parameters and generates a human mesh by decoding them~\cite{bogo2016keep,lassner2017up,pavlakos2018l3d,omran2018neural,kanazawa2018hmr,kolotouros2019spin,panteleris2018using,baek2019pushing,boukhayma20193d}.
On the contrary, the model-free approach regresses the coordinates of a 3D human mesh directly~\cite{kolotouros2019cmr,ge2019handgcn}.
Both approaches compute the 3D human pose by multiplying the output mesh with a joint regression matrix, which is defined in the human mesh models~\cite{loper2015smpl,romero2017mano}.

Although the recent deep learning-based methods have shown significant improvement, they have two major drawbacks.
First, when tested on in-the-wild data, the methods inherently suffer from the appearance domain gap between controlled and in-the-wild environment data.
The data captured from the controlled environments~\cite{ionescu2014human3,Joo2015panoptic} is valuable train data in 3D human pose and estimation, because it has accurate 3D annotations.
However, due to the significant difference of image appearance between the two domains, such as backgrounds and clothes, an image-based approach cannot fully benefit from the data.
The second drawback is that the pose parameters of the human mesh models might not be an appropriate regression target, as addressed in Kolotouros~et al.~\cite{kolotouros2019cmr}.  
The SMPL pose parameters, for example, represent 3D rotations in an axis-angle, which can suffer from the non-unique problem (\textit{i.e.}, periodicity).
While many works~\cite{kanazawa2018hmr,omran2018neural,lassner2017up} tried to avoid the periodicity by using a rotation matrix as the prediction target, it still has a non-minimal representation issue.

\begin{figure}[t]
\setlength\belowcaptionskip{-5ex}
\centerline{
\includegraphics[width=1.0\linewidth,trim={4pt 4pt 4pt 4pt}]{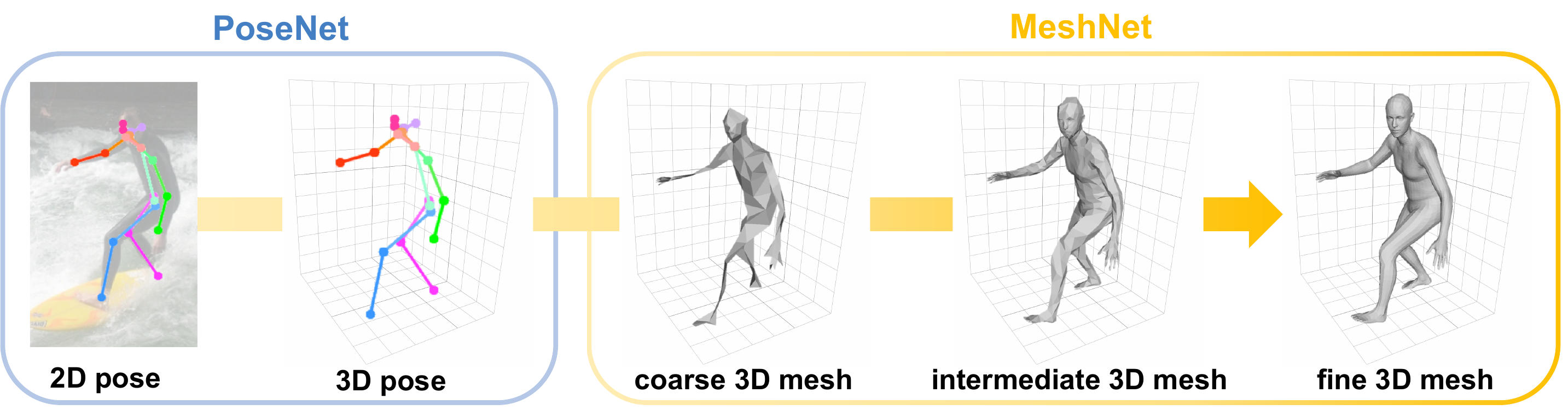}}
\caption
{
The overall pipeline of Pose2Mesh.
}
\label{fig:pipeline}
\end{figure}

To resolve the above issues, we propose Pose2Mesh, a graph convolutional system that recovers 3D human pose and mesh from the 2D human pose, in a model-free fashion.
It has two advantages over existing methods.
First, the proposed system benefits from a relatively homogeneous geometric property of the input 2D poses from controlled and in-the-wild environments.
They not only alleviates the appearance domain gap issue, but also provide essential geometric information on the human articulation.
Also, the 2D poses can be estimated accurately from in-the-wild images, since many well-performing methods~\cite{chen2018cpn,xiao2018simple,Moon_2019_CVPR_PoseFix,sun2019deep} are trained on large-scale in-the-wild 2D human pose datasets~\cite{lin2014mscoco,andriluka2018posetrack}.
The second advantage is that Pose2Mesh avoids the representation issues of the pose parameters, while exploiting the human mesh topology (\textit{i.e.}, face and edge information).
It directly regresses the 3D coordinates of mesh vertices using a graph convolutional neural network (GraphCNN) with graphs constructed from the mesh topology.

We designed Pose2Mesh in a cascaded architecture, which consists of PoseNet and MeshNet.
PoseNet lifts the 2D human pose to the 3D human pose.
MeshNet takes both 2D and 3D human poses to estimate the 3D human mesh in a coarse-to-fine manner.
During the forward propagation, the mesh features are initially processed in a coarse resolution and gradually upsampled to a fine resolution.
Figure~\ref{fig:pipeline} depicts the overall pipeline of the system.

The experimental results show that the proposed Pose2Mesh outperforms the previous state-of-the-art 3D human pose and mesh estimation methods~\cite{kolotouros2019spin,kanazawa2018hmr,kolotouros2019cmr} on various publicly available 3D human body and hand datasets~\cite{von20183dpw,ionescu2014human3,chris2019Freihand}.
Particularly, our Pose2Mesh provides the state-of-the-art result on in-the-wild dataset~\cite{von20183dpw}, even when it is trained only on the controlled setting dataset~\cite{ionescu2014human3}.

We summarize our contributions as follows. 
    \begin{itemize}
        \item We propose a novel system, Pose2Mesh, that recovers 3D human pose and mesh from the 2D human pose.
        The input 2D human pose lets Pose2Mesh robust to the appearance domain gap between controlled and in-the-wild environment data.

        \item Our Pose2Mesh directly regresses 3D coordinates of a human mesh using GraphCNN.
        It avoids representation issues of the model parameters and leverages the pre-defined mesh topology.
        
        \item We show that Pose2Mesh outperforms previous 3D human pose and mesh estimation methods on various publicly available datasets.
    \end{itemize}

\vspace{-8mm}
\section{Related works}
\vspace{-1mm}
\noindent\textbf{3D human body pose estimation.}
Current 3D human body pose estimation methods can be categorized into two approaches according to the input type: an image-based approach and a 2D pose-based approach.
The image-based approach takes an RGB image as an input for 3D body pose estimation.
Sun~et al.~\cite{sun2017chp} proposed to use compositional loss, which exploits the joint connection structure.
Sun~et al.~\cite{sun2018integral} employed soft-argmax operation to regress the 3D coordinates of body joints in a differentiable way.
Sharma~et al.~\cite{sharma2019ordinal} incorporated a generative model and depth ordering of joints to predict the most reliable 3D pose that corresponds to the estimated 2D pose.

The 2D pose-based approach lifts the 2D human pose to the 3D space.
Martinez~et al.~\cite{martinez2017simpyet} introduced a simple network that consists of consecutive fully-connected layers, which lifts the 2D human pose to the 3D space.
Zhao~et al.~\cite{zhao2019semgcn} developed a semantic GraphCNN to use spatial relationships between joint coordinates. 
Our work follows the 2D pose-based approach, to make the Pose2Mesh more robust to the domain difference between the controlled environment of the training set and in-the-wild environment of the testing set. 

\noindent\textbf{3D human body and hand pose and mesh estimation.}
A model-based approach trains a neural network to estimate the human mesh model parameters~\cite{loper2015smpl,romero2017mano}.
It has been widely used for the 3D human mesh estimation, since it does not necessarily require 3D annotation for mesh supervision.
Pavlakos~et al.~\cite{pavlakos2018l3d} proposed a system that could be only supervised by 2D joint coordinates and silhouette. 
Omran~et al.~\cite{omran2018neural} trained a network with 2D joint coordinates, which takes human part segmentation as input.
Kanazawa~et al.~\cite{kanazawa2018hmr} utilized adversarial loss to regress plausible SMPL parameters.
Baek~et al.~\cite{baek2019pushing} trained a CNN to estimate parameters of the MANO model using neural renderer~\cite{kato2018neural}.
Kolotouros~et al.~\cite{kolotouros2019spin} introduced a self-improving system that consists of SMPL parameter regressor and iterative fitting framework~\cite{bogo2016keep}. 

Recently, the advance of fitting frameworks~\cite{bogo2016keep,pavlakos2019expressive} has motivated a model-free approach, which estimates human mesh coordinates directly.
It enabled researchers to obtain 3D mesh annotation, which is essential for the model-free methods, from in-the-wild data. 
Kolotouros~et al.~\cite{kolotouros2019cmr} proposed a GraphCNN, which learns the deformation of the template body mesh to the target body mesh. 
Ge~et al.~\cite{ge2019handgcn} adopted a GraphCNN to estimate vertices of hand mesh.
Moon~et al.~\cite{moon2020i2l} proposed a new heatmap representation, called lixel, to recover 3D human meshes.

Our Pose2Mesh differs from the above methods, which are image-based, in that it uses the 2D human pose as an input.
The proposed system can benefit from the data with 3D annotations, which are captured from
controlled environments~\cite{ionescu2014human3,Joo2015panoptic}, without the appearance domain gap issue.

\noindent\textbf{GraphCNN for mesh processing.}
Recently, many methods consider a mesh as a graph structure and process it using the GraphCNN, since it can fully exploit mesh topology compared with simple stacked fully-connected layers.
Wang~et al.~\cite{wang2018pix2mesh} adopted a GraphCNN to learn a deformation from an initial ellipsoid mesh to the target object mesh in a coarse-to-fine manner.
Verma~et al.~\cite{verma2018featStNet} proposed a novel graph convolution operator for the shape correspondence problem.
Ranjan~et al.~\cite{ranjan2018comma} also proposed a GraphCNN-based VAE, which learns a latent space of the human face meshes in a hierarchical manner.

\vspace{-3mm}
\section{PoseNet}~\label{sec:posenet}
\vspace{-12mm}
\subsection{Synthesizing errors on the input 2D pose}
PoseNet estimates the root joint-relative 3D pose $\mathbf{P}^{\text{3D}} \in \mathbb{R}^{J \times 3}$ from the 2D pose, where $J$ denotes the number of human joints.
We define the root joint of the human body and hand as pelvis and wrist, respectively.
The estimated 2D pose often contains errors~\cite{ruggero2017benchmarking}, especially under severe occlusions or challenging poses.
To make PoseNet robust to the errors, we synthesize 2D input poses by adding realistic errors on the ground truth 2D pose, following ~\cite{Moon_2019_CVPR_PoseFix,chang2020abs}, during the training stage.
We represent the estimated 2D pose or the synthesized 2D pose as $\mathbf{P}^{\text{2D}} \in \mathbb{R}^{J \times 2}$.

\vspace{-3mm}
\subsection{2D input pose normalization}
We apply standard normalization to $\mathbf{P}^{\text{2D}}$, following~\cite{wandt2019repnet,chang2020abs}.
For this, we subtract the mean from $\mathbf{P}^{\text{2D}}$ and divide it by the standard deviation, which becomes $\bar{\mathbf{P}}^{\text{2D}}$.
The mean and the standard deviation of $\mathbf{P}^{\text{2D}}$ represent the 
2D location and scale of the subject, respectively.
This normalization is necessary because $\mathbf{P}^{\text{3D}}$ is independent of  scale and location of the 2D input pose $\mathbf{P}^{\text{2D}}$.

\vspace{-3mm}
\subsection{Network architecture}
The architecture of the PoseNet is based on that of ~\cite{martinez2017simpyet,chang2020abs}.
The normalized 2D input pose $\bar{{\mathbf{P}}}^{\text{2D}}$ is converted to a 4096-dimensional feature vector through a fully-connected layer.
Then, it is fed to the two residual blocks~\cite{he2016res}.
Finally, the output feature vector of the residual blocks is converted to $(3J)$-dimensional vector, which represents $\mathbf{P}^{\text{3D}}$, by a full-connected layer.

\subsection{Loss function}
We train the PoseNet by minimizing $L1$ distance between the predicted 3D pose $\mathbf{P}^{\text{3D}}$ and groundtruth.
The loss function $L_{\text{pose}}$ is defined as follows:
\begin{equation}
L_{\text{pose}} = \|\mathbf{P}^{\text{3D}} - {\mathbf{P}^{\text{3D}}}^{*}\|_1,
\end{equation}
where the asterisk indicates the groundtruth.
\begin{figure}[t]
\centerline{
\includegraphics[width=1.0\linewidth,trim={4pt 4pt 4pt 4pt}]{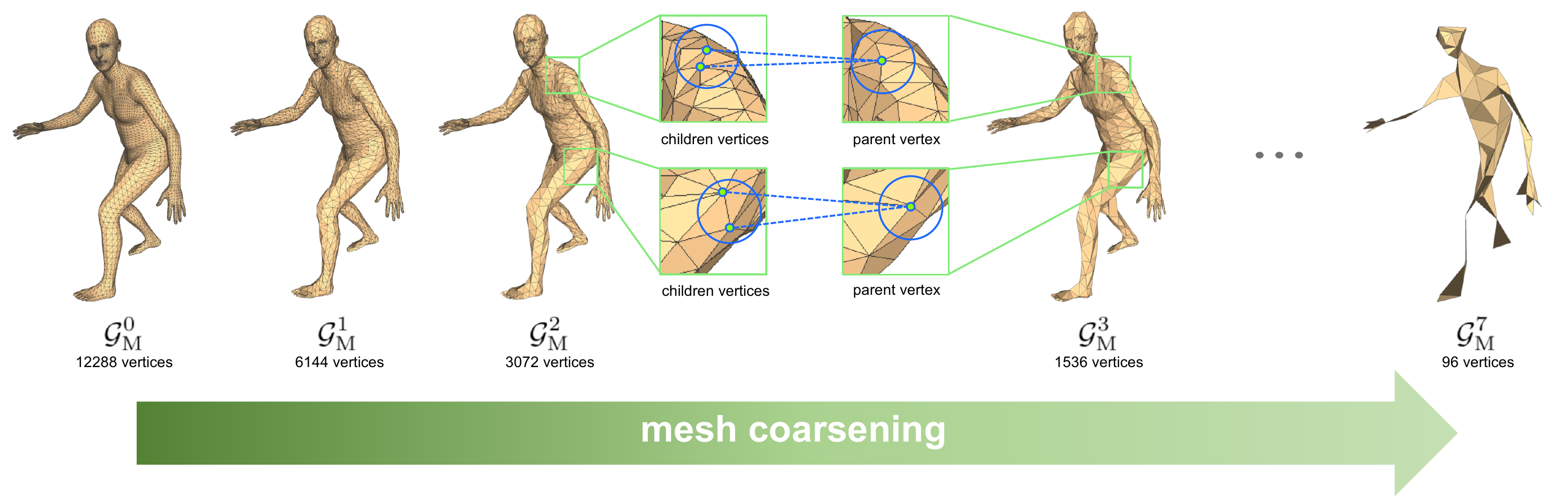}}
\caption
{
The coarsening initially generates multiple coarse graphs from $\mathcal{G}_\text{M}$, and adds fake nodes without edges to each graph, following~\cite{defferrard2016chebygcn}.
The numbers of vertices range from 96 to 12288 for body meshes and from 68 to 1088 for hand meshes.
}
\label{fig:coarsening}
\end{figure}

\begin{figure}[t]
\setlength\belowcaptionskip{-4ex}
\centerline{
\includegraphics[width=0.8\linewidth,trim={4pt 4pt 4pt 4pt}]{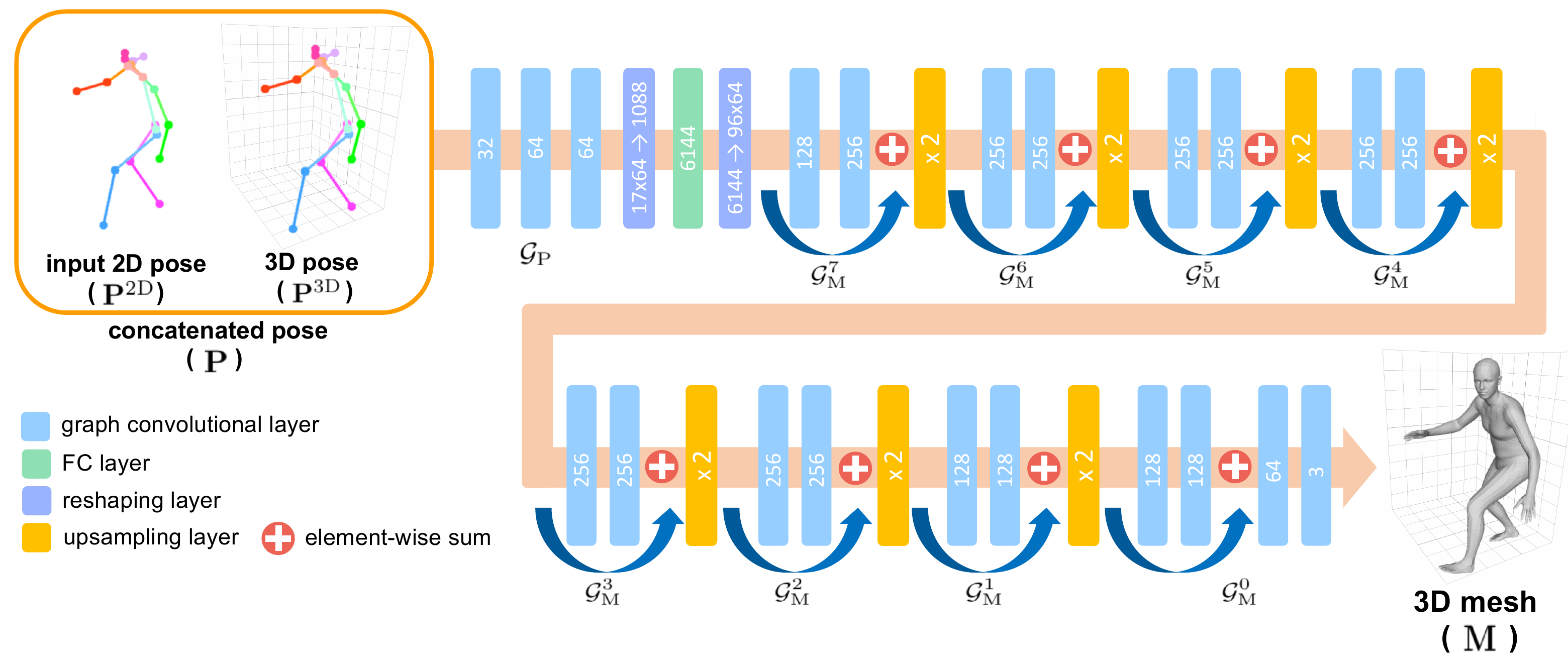}}
\caption
{
The network architecture of MeshNet.
}
\label{fig:meshnet_architecture}
\end{figure}

\vspace{-2mm}
\section{MeshNet}
\vspace{-2mm}

\subsection{Graph convolution on pose}
MeshNet concatenates $\bar{\mathbf{P}}^{\text{2D}}$ and $\mathbf{P}^{\text{3D}}$ into $\mathbf{P} \in \mathbb{R}^{J \times 5}$.
Then, it estimates the root joint-relative 3D mesh $\mathbf{M} \in \mathbb{R}^{V \times 3}$ from $\mathbf{P}$, where $V$ denotes the number of human mesh vertices.
To this end, MeshNet uses the spectral graph convolution~\cite{shuman2013gsp,joan2014spect}, which can be defined as the multiplication of a signal $x \in \mathbb{R}^N$ with a filter $g_\theta=\textit{diag}(\theta)$ in Fourier domain as follows:
\begin{equation}
g_\theta * x = \mathit{U}g_\theta \mathit{U}^T x,
\end{equation}
where graph Fourier basis $\mathit{U}$ is the matrix of the eigenvectors of the normalized graph Laplacian $\mathit{L}$~\cite{chung1997spectgraph}, and $\mathit{U}^Tx$ denotes the graph Fourier transform of $x$.
Specifically, to reduce the computational complexity, we design MeshNet to be based on Chebysev spectral graph convolution~\cite{defferrard2016chebygcn}.

\smallbreak
\noindent \textbf{Graph construction.}
We construct a graph of $\mathbf{P}$,  $\mathcal{G}_\text{P}=(\mathcal{V}_\text{P},\mathit{A}_\text{P})$, where $\mathcal{V}_\text{P}=\mathbf{P}=\{\mathbf{p}_i\}^{J}_{i=1}$ is a set of $J$ human joints, and $\mathit{A}_\text{P} \in \{0,1\}^{J \times J}$ is an adjacency matrix.
$\mathit{A}_\text{P}$ defines the edge connections between the joints based on the human skeleton and symmetrical relationships~\cite{cai2019temp_gcn}, where $(\mathit{A}_\text{P})_{ij}=1$ if joints $i$ and $j$ are the same or connected, and $(\mathit{A}_\text{P})_{ij}=0$ otherwise.
The normalized Laplaican is computed as $\mathit{L}_{\text{P}}=\mathit{I}_J - \mathit{D}^{-1/2}_{\text{P}}\mathit{A}_{\text{P}}\mathit{D}^{-1/2}_{\text{P}}$, where $\mathit{I}_J$ is the identity matrix, and $\mathit{D}_{\text{P}}$ is the diagonal matrix which represents the degree of each joint in $\mathcal{V}_\text{P}$ as $(\mathit{D}_{\text{P}})_{ij}=\sum_j (\mathit{A}_\text{P})_{ij}$.
The scaled Laplacian is computed as $\tilde{\mathit{L}_{\text{P}}}=2\mathit{L}_{\text{P}}/\lambda_{\text{max}}-\mathit{I}_J$.

\smallbreak
\noindent \textbf{Spectral convolution on graph.}
Then, MeshNet performs the spectral graph convolution on $\mathcal{G}_\text{P}$, which is defined as follows:

\begin{equation}\label{eq:2}
\mathit{F}_{\text{out}} = \sum_{k=0}^{K-1}\mathit{T}_k \big(\tilde{\mathit{L}_{\text{P}}}\big)\mathit{F}_{\text{in}}\mathit{\Theta}_k,       
\end{equation}
where $\mathit{F}_{\text{in}} \in \mathbb{R}^{J \times f_{
\text{in}}}$ and $\mathit{F}_{\text{out}} \in \mathbb{R}^{J \times f_{\text{out}}}$ are the input and output feature maps respectively, $\mathit{T}_k \big(x\big)=2x\mathit{T}_{k-1} \big(x\big)-\mathit{T}_{k-2} \big(x\big)$ is the Chebysev polynomial~\cite{hammond2009wavelet} of order $k$, and $\mathit{\Theta}_k \in \mathbb{R}^{f_{\text{in}} \times f_{\text{out}}}$ is the $k$th Chebysev coefficient matrix, whose elements are the trainable parameters of the graph convolutional layer.
$f_{\text{in}}$ and $f_{\text{out}}$ are the input and output feature dimensions respectively.
The initial input feature map  $\mathit{F}_{\text{in}}$ is $\mathbf{P}$ in practice, where $f_{\text{in}}=5$.
This graph convolution is $K$-localized, which means at most $K$-hop neighbor nodes from each node are affected~\cite{defferrard2016chebygcn,kipf2017gcn}, since it is a $K$-order polynomial in the Laplacian.
Our MeshNet sets $K=3$ for all graph convolutional layers following~\cite{ge2019handgcn}.

\vspace{-2mm}
\subsection{Coarse-to-fine mesh upsampling}
We gradually upsample $\mathcal{G}_\text{P}$ to the graph of $\mathbf{M}$,
$\mathcal{G}_\text{M}=(\mathcal{V}_\text{M},\mathit{A}_\text{M})$, where $\mathcal{V}_\text{M}=\mathbf{M}=\{\mathbf{m}_i\}^{V}_{i=1}$ is a set of $V$ human mesh vertices, and $\mathit{A}_\text{M} \in \{0,1\}^{V \times V}$ is an adjacency matrix defining edges of the human mesh.
To this end, we apply the graph coarsening~\cite{dhillon2007graculus} technique to $\mathcal{G}_\text{M}$, which creates various resolutions of graphs, $\{\mathcal{G}_\text{M}^c=(\mathcal{V}_\text{M}^c,\mathit{A}_\text{M}^c)\}_{c=0}^C$, where $C$ denotes the number of coarsening steps, following Defferrard~et al.~\cite{defferrard2016chebygcn}.
Figure~\ref{fig:coarsening} shows the coarsening process and a balanced binary tree structure of mesh graphs, where the $i$th vertex in $\mathcal{G}_\text{M}^{c+1}$ is a parent node of the $2i-1$th and $2i$th vertices in $\mathcal{G}_\text{M}^{c}$, and  $2|\mathcal{V}_\text{M}^{c+1}|=|\mathcal{V}_\text{M}^c|$. 
$i$ starts from 1.
The final output of MeshNet is $\mathcal{V}_\text{M}$, which is converted from  $\mathcal{V}_\text{M}^0$ by a pre-defined indices mapping.
During the forward propagation, MeshNet first upsamples the $\mathcal{G}_\text{P}$ to the coarsest mesh graph $\mathcal{G}_\text{M}^C$ by reshaping and a fully-connected layer.
Then, it performs the spectral graph convolution on each resolution of mesh graphs as follows:
\begin{equation}
\mathit{F}_{\text{out}} = \sum_{k=0}^{K-1}\mathit{T}_k \big(\tilde{\mathit{L}_{\text{M}}^c}\big)\mathit{F}_{\text{in}}\mathit{\Theta}_k,  
\end{equation}
where $\tilde{\mathit{L}_{\text{M}}^c}$ denotes the scaled Laplacian of $\mathcal{G}_\text{M}^c$, and the other notations are defined in the same manner as Equation~\ref{eq:2}. 
Following~\cite{ge2019handgcn}, MeshNet performs mesh upsampling by copying features of each parent vertex in $\mathcal{G}_\text{M}^{c+1}$ to the corresponding children vertices in $\mathcal{G}_\text{M}^c$.
The upsampling process is defined as follows:
\begin{equation}
\mathit{F}_c = \psi(\mathit{F}_{c+1}^T)^T,
\end{equation}
where $\mathit{F}_{c} \in \mathbb{R}^{\mathcal{V}_\text{M}^{c} \times f_{c}}$ is the first feature map of $\mathcal{G}_\text{M}^{c}$, $\mathit{F}_{c+1} \in \mathbb{R}^{\mathcal{V}_\text{M}^{c+1} \times f_{c+1}}$ is the last feature map of $\mathcal{G}_\text{M}^{c+1}$, $\psi \colon \mathbb{R}^{f_{c+1} \times \mathcal{V}_\text{M}^{c+1}} \to \mathbb{R}^{f_{c+1} \times \mathcal{V}_\text{M}^c}$ denotes a nearest-neighbor upsampling function, and $f_{c}$ and $f_{c+1}$are the feature dimensions of vertices in $\mathit{F}_{c}$ and $\mathit{F}_{c+1}$ respectively.
The nearest upsampling function copies the feature of the $i$th vertex in $\mathcal{G}_\text{M}^{c+1}$ to the $2i-1$th and $2i$th vertices in  $\mathcal{G}_\text{M}^c$.
To facilitate the learning process, we additionally incorporate a residual connection between each resolution.
Figure~\ref{fig:meshnet_architecture} shows the overall architecture of MeshNet.

\vspace{-2mm}
\subsection{Loss functions}
To train our MeshNet, we use four loss functions.

\noindent\textbf{Vertex coordinate loss.}
We minimize $L1$ distance between the predicted 3D mesh coordinates $\mathbf{M}$ and groundtruth, which is defined as follows:
\begin{equation}
L_{\text{vertex}} = \|\mathbf{M} - \mathbf{M}^{*}\|_1,
\end{equation}
where the asterisk indicates the groundtruth.

\noindent\textbf{Joint coordinate loss.}
We use a $L1$ loss function between the groundtruth root-relative 3d pose and the 3D pose regressed from $\mathbf{M}$, to train our MeshNet to estimate mesh vertices aligned with joint locations.
The 3D pose is calculated as $\mathcal{J}\mathbf{M}$, where $\mathcal{J} \in \mathbb{R}^{J \times V}$ is a joint regression matrix defined in SMPL or MANO model. 
The loss function is defined as follows:
\begin{equation}
L_{\text{joint}} = \|\mathcal{J}\mathbf{M} - {\mathbf{P}^{\text{3D}}}^{*}\|_1,
\end{equation}
where the asterisk indicates the groundtruth.

\noindent\textbf{Surface normal loss.}
We supervise normal vectors of an output mesh surface to be consistent with groundtruth.
This consistency loss improves surface smoothness and local details~\cite{wang2018pix2mesh}.
The loss function $L_{\text{normal}}$ is defined as follows:
\begin{equation}
L_{\text{normal}} = \sum_{f} \sum_{\{i,j\} \subset f} \Big |\Big \langle \frac{\mathbf{m}_{i} - \mathbf{m}_{j}} {\|\mathbf{m}_{i} - \mathbf{m}_{j}\|_2} ,  n^*_f \Big \rangle \Big |,
\end{equation}
where $f$ and $n^*_f$ denote a triangle face in the human mesh and a groundtruth unit normal vector of $f$, respectively. $\mathbf{m}_i$ and $\mathbf{m}_j$ denote the $i$th and $j$th vertices in $f$.

\noindent\textbf{Surface edge loss.}
We define edge length consistency loss between predicted and groundtruth edges, following~\cite{wang2018pix2mesh}.
The edge loss is effective in recovering smoothness of hands, feet, and a mouth, which have dense vertices.
The loss function $L_{\text{edge}}$ is defined as follows:
\begin{equation}
L_{\text{edge}} = \sum_{f} \sum_{\{i,j\} \subset f} | \|\mathbf{m}_{i} - \mathbf{m}_{j}\|_2 - \|\mathbf{m}^{*}_{i} - \mathbf{m}^{*}_{j}\|_2 |,
\end{equation}
where $f$ and the asterisk denote a triangle face in the human mesh and the groundtruth, respectively.
$\mathbf{m}_i$ and $\mathbf{m}_j$ denote $i$th and $j$th vertex in $f$.

We define the total loss of our MeshNet, $L_{\text{mesh}}$, as a weighted sum of all four loss functions:
\begin{equation}
L_{\text{mesh}} = \lambda_\text v L_{\text{vertex}} + \lambda_\text j L_{\text{joint}} + \lambda_\text n L_{\text{normal}} + \lambda_\text e L_{\text{edge}}, 
\end{equation}
where $\lambda_\text v=1,\;\lambda_\text j=1,\;\lambda_\text n=0.1,$ and $\lambda_\text e=20$.

\vspace{-1mm}
\section{Implementation Details}
\vspace{-1mm}
PyTorch~\cite{paszke2017automatic} is used for implementation. 
We first pre-train our PoseNet, and then train the whole network, Pose2Mesh, in an end-to-end manner.
Empirically, our two-step training strategy gives better performance than the one-step training.
The weights are updated by the Rmsprop optimization~\cite{hinton2012rmsprop} with a mini-batch size of 64.
We pre-train PoseNet 60 epochs with a learning rate $10^{-3}$.
The learning rate is reduced by a factor of 10 after the $30$th epoch.
After integrating the pre-trained PoseNet to Pose2Mesh, we train the whole network 15 epochs with a learning rate $10^{-3}$.  
The learning rate is reduced by a factor of 10 after the $12$th epoch.
In addition, we set $\lambda_\text e$ to 0 until 7 epoch on the second training stage, since it tends to cause local optima at the early training phase.
We used four NVIDIA RTX 2080 Ti GPUs for Pose2Mesh training, which took at least a half day and at most two and a half days, depending on the training datasets.
In inference time, we use 2D pose outputs from Sun~et al.~\cite{sun2019deep} and Xiao~et al.~\cite{xiao2018simple}.
They run at 5 fps and 67 fps respectively, and our Pose2Mesh runs at 37 fps.
Thus, the proposed system can process from 4 fps to 22 fps in practice, which shows the applicability to real-time applications. 

\vspace{-2mm}
\section{Experiment}
\vspace{-1mm}
\subsection{Datasets and evaluation metrics}
\textbf{Human3.6M.}
Human3.6M~\cite{ionescu2014human3} is a large-scale indoor 3D body pose benchmark, which consists of 3.6M video frames. 
The groundtruth 3D poses are obtained using a motion capture system, but there are no groundtruth 3D meshes.
As a result, for 3D mesh supervision, most of the previous 3D pose and mesh estimation works~\cite{kanazawa2018hmr,kolotouros2019cmr,kolotouros2019spin} used pseudo-groundtruth obtained from Mosh~\cite{loper2014mosh}.
However, due to the license issue, the pseudo-groundtruth from Mosh is not currently publicly accessible.
Thus, we generate new pseudo-groundtruth 3D meshes by fitting SMPL parameters to the 3D groundtruth poses using SMPLify-X~\cite{pavlakos2019expressive}.
For the fair comparison, we trained and tested previous state-of-the-art methods on the obtained groundtruth using their officially released code.
Following~\cite{pavlakos2017cf,kanazawa2018hmr}, all methods are trained on 5 subjects (S1, S5, S6, S7, S8) and tested on 2 subjects (S9, S11).

We report our performance for the 3D pose using two evaluation metrics.
One is mean per joint position error (MPJPE)~\cite{ionescu2014human3}, which measures the Euclidean distance in millimeters between the estimated and groundtruth joint coordinates, after aligning the root joint.
The other one is PA-MPJPE, which calculates MPJPE after further alignment (\textit{i.e.}, Procrustes analysis (PA)~\cite{gower1975generalized}).
$\mathcal{J}\mathbf{M}$ is used for the estimated joint coordinates.
We only evaluate 14 joints out of 17 estimated joints following ~\cite{kanazawa2018hmr,pavlakos2018l3d,kolotouros2019cmr,kolotouros2019spin}.

\noindent\textbf{3DPW.}
3DPW~\cite{von20183dpw} is captured from in-the-wild and contains 3D body pose and mesh annotations.
It consists of 51K video frames, and IMU sensors are leveraged to acquire the groundtruth 3D pose and mesh.
We only use the test set of 3DPW for evaluation following~\cite{kolotouros2019spin}.
MPJPE and mean per vertex position error (MPVPE) are used for evaluation.
14 joints from $\mathcal{J}\mathbf{M}$, whose joint set follows that of Human3.6M, are evaluated for MPJPE as above.
MPVPE measures the Euclidean distance in millimeters between the estimated and groundtruth vertex coordinates, after aligning the root joint.

\noindent\textbf{COCO.}
COCO~\cite{lin2014mscoco} is an in-the-wild dataset with various 2D annotations such as detection and human joints.
To exploit this dataset on 3D mesh learning, Kolotouros~et al.~\cite{kolotouros2019spin} fitted SMPL parameters to 2D joints using SMPLify~\cite{bogo2016keep}.
Following them, we use the processed data for training.

\noindent \textbf{MuCo-3DHP.}
MuCo-3DHP~\cite{mehta2018muco} is synthesized from the existing MPI-INF-3DHP 3D single-person pose estimation dataset~\cite{mehta2017monocular}. 
It consists of 200K frames, and half of them have augmented backgrounds.
For the background augmentation, we use images of COCO that do not include humans to follow Moon~et al.~\cite{moon2019camera}.
Following them, we use this dataset only for the training.

\noindent\textbf{FreiHAND.}
FreiHAND~\cite{chris2019Freihand} is a large-scale 3D hand pose and mesh dataset.
It consists of a total of 134K frames for training and testing.
Following Zimmermann~et al.~\cite{chris2019Freihand}, we report PA-MPVPE, F-scores, and additionally PA-MPJPE of Pose2Mesh.
$\mathcal{J}\mathbf{M}$ is evaluated for the joint errors.

\begin{table}[t]
\setlength{\tabcolsep}{1pt}
\setlength{\belowcaptionskip}{3pt plus 3pt minus 2pt}
\centering
\resizebox{1.0\textwidth}{!}{\begin{minipage}{\textwidth}
\caption{The performance comparison between four combinations of regression target and network design tested on Human3.6M. `no. param.' denotes the number of parameters of a network, which estimates SMPL parameters or vertex coordinates from the output of PoseNet.} 
\centering
\scalebox{1.0}{
\begin{tabular}{C{2.3cm}|C{1.2cm}C{1.7cm}C{1.7cm}|C{1.2cm}C{1.7cm}C{1.7cm}}
\specialrule{.1em}{.05em}{.05em}
\small\multirow{ 2}{*}{ target$\backslash$network}
 & \multicolumn{3}{c|}{ FC} & \multicolumn{3}{c}{ GraphCNN} \\ \cline{2-7}
                       &  MPJPE &  PA-MPJPE &  no. param. &
                        MPJPE &  PA-MPJPE &  no. param. \\ \hline
 SMPL param. &  72.8 &  55.5 &  17.3M &  79.1 &  59.1 &  13.5M \\
 vertex coord. &  119.6 &  95.1 &  37.5M &  \textbf{64.9} &  \textbf{48.0} &  \textbf{8.8M} \\
 \specialrule{.1em}{.05em}{.05em}
\end{tabular}
}
\label{table:output_network}
\end{minipage}}
\end{table}
\vspace{-2mm}

\begin{table}[t]
\setlength{\tabcolsep}{1.0pt}
\def\arraystretch{1.1}
\centering
\resizebox{1.0\textwidth}{!}{\begin{minipage}{\textwidth}
\begin{minipage}{.51\linewidth}
\caption{The performance comparison on Human3.6M between two upsampling schems. GPU mem. and fps denote the required memory during training and fps in inference time respectively.}
\centering
\scalebox{1.0}{
\begin{tabular}{C{2.2cm}|C{1.7cm}C{0.5cm}C{1.2cm}C{1.7cm}}
\specialrule{.1em}{.05em}{.05em}
 method & GPU mem. & fps & MPJPE  \\ \hline   
 direct & 10G & 24 & 65.3 \\
 \textbf{coarse-to-fine} & \textbf{6G} &  \textbf{37} & \textbf{64.9} \\
 \specialrule{.1em}{.05em}{.05em}
\end{tabular}
}
\label{table:coarse-to-fine-metric}
\end{minipage}\hfill
\begin{minipage}{.45\linewidth}
\centering
\setlength{\belowcaptionskip}{8pt plus 3pt minus 2pt}
\caption{The MPJPE comparison between four architectures tested on 3DPW.}
\scalebox{1.0}{
\begin{tabular}{C{3.5cm}|C{1.4cm}C{1.4cm}}
\specialrule{.1em}{.05em}{.05em}
 architecture &  MPJPE \\ \hline
 2D$\rightarrow$mesh &  101.1  \\
 2D$\rightarrow$3D$\rightarrow$mesh &  103.2 \\
 \textbf{2D$\rightarrow$3D+2D$\rightarrow$mesh} &  \textbf{100.5} \\
 \specialrule{.1em}{.05em}{.05em}
\end{tabular}
}
\label{table:cascaded_architecture}
\end{minipage}
\end{minipage}}
\end{table}

\begin{table}[t]
\setlength{\tabcolsep}{1pt}
\def\arraystretch{1.2}
\centering
\resizebox{1.0\textwidth}{!}{\begin{minipage}{\textwidth}
\caption{The upper bounds of the two different graph convolutional networks that take a 2D pose and a 3D pose. Tested on Human3.6M.}
\centering
\scalebox{1.0}{
\begin{tabular}{C{2.8cm}|C{2.5cm}|C{1.5cm}C{2.4cm}}
\specialrule{.1em}{.05em}{.05em}
 test input &  architecture &  MPJPE &  PA-MPJPE \\ \hline
 2D pose GT &  2D$\rightarrow$mesh &  55.5 &  38.4 \\ \hline
 3D pose from~\cite{moon2019camera} &  3D$\rightarrow$mesh &  56.3 &  43.2 \\
 \textbf{3D pose GT} &  \textbf{3D$\rightarrow$mesh} &  \textbf{29.0} &  \textbf{23.0} \\
 \specialrule{.1em}{.05em}{.05em}
\end{tabular}
}
\label{table:upper_bound}
\end{minipage}}
\end{table}
\vspace{-2mm}

\subsection{Ablation study}
To analyze each component of the proposed system, we trained different networks on Human3.6M, and evaluated on Human3.6M and 3DPW.
The test 2D input poses used in Human3.6M and 3DPW evaluation are outputs from Integral Regression~\cite{sun2018integral} and HRNet~\cite{sun2019deep} respectively, which are obtained using groundtruth bounding boxes.

\noindent\textbf{Regression target and network design.}
To demonstrate the effectiveness of regressing the 3D mesh vertex coordinates using GraphCNN, we compare MPJPE and PA-MPJPE of four different combinations of the regression target and the network design in Table~\ref{table:output_network}.
First, \textit{vertex-GraphCNN}, our Pose2Mesh, substantially improves the joint errors compared to \textit{vertex-FC}, which regresses vertex coordinates with a network of fully-connected layers.
This proves the importance of exploiting the human mesh topology with GraphCNN, when estimating the 3D vertex coordinates.
Second, \textit{vertex-GraphCNN} provides better performance than both networks estimating SMPL parameters, while maintaining the considerably smaller number of network parameters.
Taken together, the effectiveness of our mesh coordinate regression scheme using GraphCNN is clearly justified.

In this comparison, the same PoseNet and cascaded architecture are employed for all networks.
On top of the PoseNet, \textit{vertex-FC} and \textit{param-FC} used a series of fully-connected layers, whereas \textit{param-GraphCNN} added fully-connected layers on top of Pose2Mesh.
For the fair comparison, when training \textit{param-FC} and \textit{param-GraphCNN}, we also supervised the reconstructed mesh from the predicted SMPL parameters with $L_{\text{vertex}}$ and $L_{\text{joint}}$.
The networks estimating SMPL parameters incorporated Zhou~et al.'s method~\cite{zhou2019continuity} for continuous rotations.

\noindent\textbf{Coarse-to-fine mesh upsampling.}
We compare a coarse-to-fine mesh upsampling scheme and a direct mesh upsampling scheme.
The direct upsampling method performs graph convolution on the lowest resolution mesh until the middle layer of MeshNet, and then directly upsamples it to the highest one (\textit{e.g.}, 96 to 12288 for the human body mesh).
While it has the same number of graph convolution layers and almost the same number of parameters, our coarse-to-fine model consumes half as much GPU memory and runs 1.5 times faster than the direct upsampling method.
It is because graph convolution on the highest resolution takes much more time and memory than graph convolution on lower resolutions.
In addition, the coarse-to-fine upsampling method provides a slightly lower joint error, as shown in Table~\ref{table:coarse-to-fine-metric}.
These results confirm the effectiveness of our coarse-to-fine upsampling strategy.

\noindent\textbf{Cascaded architecture analysis.}
We analyze the cascaded architecture of Pose2Mesh to demonstrate its validity in Table~\ref{table:cascaded_architecture}.
To be specific, we construct (a) a GraphCNN that directly takes a 2D pose, (b) a cascaded network that predicts mesh coordinates from a 3D pose from pretrained PoseNet, and (c) our Pose2Mesh.
All methods are both trained by synthesized 2D poses.
First, (a) outperforms (b), which implies a 3D pose output from PoseNet may lack geometry information in the 2D input pose.
If we concatenate the 3D pose output with the 2D input pose as (c), it provides the lowest errors.
This explains that depth information in 3D poses could positively affect 3D mesh estimation.

To further verify the superiority of the cascaded architecture, we explore the upper bounds of (a) and (d) a GraphCNN that takes a 3D pose in Table~\ref{table:upper_bound}.
To this end, we fed the groundtruth 2D pose and 3D pose to (a) and (d) as test inputs, respectively.
Apparently, since the input 3D pose contains additional depth information, the upper bound of (d) is considerably higher than that of (a).
We also fed state-of-the-art 3D pose outputs from~\cite{moon2019camera} to (d), to validate the practical potential for performance improvement.
Surprisingly, the performance is comparable to the upper bound of (a).
Thus, our Pose2Mesh will substantially outperform (a) a graph convolution network that directly takes a 2D pose, if we can improve the performance of PoseNet. In summary, the above results prove the validity of our cascaded architecture of Pose2Mesh.

\begin{table}[t]
\setlength{\tabcolsep}{1pt}
\def\arraystretch{1.1}
\centering
\resizebox{1.0\linewidth}{!}{\begin{minipage}{\linewidth}
\centering
\setlength{\belowcaptionskip}{3pt plus 3pt minus 2pt}
\caption{The accuracy comparison between state-of-the-art methods and Pose2Mesh on Human3.6M. The dataset names on top are training sets. 
}
\scalebox{1.0}{
\begin{tabular}{C{2.8cm}|C{1.3cm}C{1.8cm}|C{1.3cm}C{1.8cm}}
\specialrule{.1em}{.05em}{.05em}
\multirow{ 2}{*}{ method} & \multicolumn{2}{c|}{ Human3.6M} & \multicolumn{2}{c}{ Human3.6M + COCO} \\
                      &  MPJPE &  PA-MPJPE &  MPJPE &  PA-MPJPE \\ \hline
 HMR~\cite{kanazawa2018hmr} &  184.7 &  88.4 &  153.2 &  85.5 \\
 GraphCMR~\cite{kolotouros2019cmr} &  148.0 &  104.6 &  78.3 &  59.5 \\
 SPIN~\cite{kolotouros2019spin} &  85.6 &  55.6 &  72.9 &  51.9 \\
 \textbf{Pose2Mesh (Ours)} &  \textbf{64.9} &  \textbf{48.0} &  \textbf{67.9} &  \textbf{49.9} \\
 \specialrule{.1em}{.05em}{.05em}
\end{tabular}
}
\label{table:compare_h36m}
\end{minipage}}
\end{table}
\vspace{-2mm}

\begin{table}[t]
\setlength{\tabcolsep}{1pt}
\def\arraystretch{1.1}
\centering
\resizebox{1.0\textwidth}{!}{\begin{minipage}{\textwidth}
\setlength{\belowcaptionskip}{3pt plus 3pt minus 2pt}
\caption{The accuracy comparison between state-of-the-art methods and Pose2Mesh on 3DPW. The dataset names on top are training sets. 
}
\centering
\scalebox{1.0}{
\begin{tabular}{C{3.1cm}|C{1.3cm}C{1.8cm}C{1.3cm}|C{1.3cm}C{1.8cm}C{1.3cm}}
\specialrule{.1em}{.05em}{.05em}
\multirow{ 2}{*}{method} & \multicolumn{3}{c|}{ Human3.6M} & \multicolumn{3}{c}{ Human3.6M + COCO} \\
                       &  MPJPE & PA-MPJPE & MPVPE &  MPJPE &  PA-MPJPE & MPVPE\\ \hline
 HMR~\cite{kanazawa2018hmr} &  377.3 & 165.7 & 481.0 & 300.4 & 137.2 & 406.8 \\
 GraphCMR~\cite{kolotouros2019cmr} &  332.5 & 177.4 & 380.8 &  126.5 & 80.1 & 144.8 \\
SPIN~\cite{kolotouros2019spin} &  313.8 & 156.0 & 344.3 &  113.1 & 71.7 & 122.8 \\
Pose2Mesh \tiny(Simple~\cite{xiao2018simple}) & 101.8 & 64.2 & 119.1 &  92.3 & 61.0 & 110.5 \\
 \textbf{Pose2Mesh \tiny (HR~\cite{sun2019deep})} &  \textbf{100.5} &  \textbf{63.0} &  \textbf{117.5} &  \textbf{91.4} &  \textbf{60.1} &  \textbf{109.3} \\
 \specialrule{.1em}{.05em}{.05em}
\end{tabular}
}
\label{table:compare_3dpw}
\end{minipage}}
\end{table}
\vspace{-2mm}

\begin{table*}[t]
\centering
\setlength\tabcolsep{1.0pt}
\def\arraystretch{1.1}
\centering
\resizebox{1.0\textwidth}{!}{\begin{minipage}{\textwidth}
\caption{
The accuracy comparison between state-of-the-art methods and Pose2Mesh on FreiHAND.
}
\centering
\scalebox{1.0}{
\begin{tabular}{C{3.0cm}|C{1.6cm}C{1.6cm}C{1.5cm}C{1.5cm}}
\specialrule{.1em}{.05em}{.05em}
 \scriptsize method &  \scriptsize PA-MPVPE &  \scriptsize PA-MPJPE &  \scriptsize F@5 mm &  \scriptsize F@15 mm \\ \hline
 \scriptsize Hasson~et al.~\cite{hasson2019learning} & \scriptsize 13.2 & \scriptsize - & \scriptsize 0.436 & \scriptsize 0.908 \\
\scriptsize Boukhayma~et al.~\cite{boukhayma20193d} & \scriptsize 13.0 & \scriptsize - & \scriptsize 0.435 & \scriptsize 0.898 \\
\scriptsize FreiHAND~\cite{chris2019Freihand} & \scriptsize 10.7 & \scriptsize - & \scriptsize\scriptsize 0.529 & \scriptsize 0.935 \\
\scriptsize \textbf{Pose2Mesh (Ours)} & \scriptsize \textbf{7.6} & \scriptsize \textbf{7.4} & \scriptsize \textbf{0.683} & \scriptsize \textbf{0.973} \\
 \specialrule{.1em}{.05em}{.05em}
\end{tabular}
}
\label{table:compare_freihand}
\end{minipage}}
\end{table*}
\vspace{-2mm}

\subsection{Comparison with state-of-the-art methods}

\textbf{Human3.6M.}
We compare our Pose2Mesh with the previous state-of-the-art 3D body pose and mesh estimation methods on Human3.6M in Table~\ref{table:compare_h36m}.
First, when we train all methods only on Human3.6M, our Pose2Mesh significantly outperforms other methods.
However, when we train the methods additionally on COCO, the performance of the previous baselines increases, but that of Pose2Mesh slightly decreases.
The performance gain of other methods is a well-known phenomenon~\cite{sun2018integral} among image-based methods, which tend to generalize better when trained with diverse images from in-the-wild.
Whereas, our Pose2Mesh does not benefit from more images in the same manner, since it only takes the 2D pose.
We analyze the reason for the performance drop is that the test set and train set of Human3.6M have similar poses, which are from the same action categories.
Thus, overfitting the network to the poses of Human3.6M can lead to better accuracy.
Nevertheless, in both cases, our Pose2Mesh outperforms the previous methods in both MPJPE and PA-MPJPE.
The test 2D input poses for Pose2Mesh are estimated by the method of Sun~et al.~\cite{sun2018integral} trained on MPII~\cite{andriluka2014mpii}, using groundtruth bounding boxes.

\noindent\textbf{3DPW.}
We compare MPJPE, PA-MPJPE, and MPVPE of our Pose2Mesh with the previous state-of-the-art 3D body pose and mesh estimation works on 3DPW, which is an in-the-wild dataset, in Table~\ref{table:compare_3dpw}.
First, when the image-based methods are trained only on Human3.6M, they give extremely high errors.
This verifies that the image-based methods suffer from the appearance domain gap between train and test data from controlled and in-the-wild environments respectively.
In fact, since Human3.6M is an indoor dataset from the controlled environment, the image appearance from it are very different from in-the-wild image appearance.
On the contrary, the 2D pose-based approach of Pose2Mesh can benefit from accurate 3D annotations of the lab-recorded 3D datasets~\cite{ionescu2014human3} without the appearance domain gap issue, utilizing the homogeneous geometric property of 2D poses from different domains.
Indeed, Pose2Mesh gives far lower errors on in-the-wild images from 3DPW, even when it is only trained on Human3.6M while other methods are additionally trained on COCO.
The experimental results suggest that a 3D pose and mesh estimation approach may not necessarily require 3D data captured from in-the-wild environments, which is extremely challenging to acquire, to give accurate predictions.
The test 2D input poses for Pose2Mesh are estimated by HRNet~\cite{sun2019deep} and Simple~\cite{xiao2018simple} trained on COCO, using groundtruth bounding boxes.
The average precision (AP) of~\cite{sun2019deep} and~\cite{xiao2018simple} are 85.1 and 82.8 on 3DPW test set, 72.1 and 70.4 on COCO validation set, respectively.

\noindent\textbf{FreiHAND.}
We present the comparison between our Pose2Mesh and other state-of-the-art 3D hand pose and mesh estimation works in Table~\ref{table:compare_freihand}.
The proposed system outperforms other methods in various metrics, including PA-MPVPE and F-scores.
The test 2D input poses for Pose2Mesh are estimated by HRNet~\cite{sun2019deep} trained on FreiHAND~\cite{chris2019Freihand}, using bounding boxes from Mask R-CNN~\cite{he2017mask} with ResNet-50 backbone~\cite{he2016res}.

\noindent\textbf{Comparison with different train sets.}
We report MPJPE and PA-MPJPE of Pose2Mesh trained on Human3.6M, COCO, and MuCo-3DHP, and other methods trained on different train sets in Table~\ref{table:compare_h36m_3dpw_different_dataset}. 
The train sets include Human3.6M, COCO, MPII~\cite{andriluka2014mpii} , LSP~\cite{johnson2010clustered}, LSP-Extended~\cite{johnson2011learning}, UP~\cite{lassner2017unite}, and MPI-INF-3DHP~\cite{mehta2017monocular}.
Each method is trained on a different subset of them. 
In the table, the errors of~\cite{kanazawa2018hmr,kolotouros2019cmr,kolotouros2019spin} decrease by a large margin compared to the errors in Table~\ref{table:compare_h36m} and~\ref{table:compare_3dpw}.
Although it shows that the image-based methods can improve the generalizability with weak-supervision on in-the-wild 2D pose datasets, Pose2Mesh still provides the lowest errors in 3DPW, which is the in-the-wild benchmark.
This suggests that tackling the domain gap issue to fully benefit from the 3D data of controlled environments is an important task to recover accurate 3D pose and mesh from in-the-wild images.
We measured the PA-MPJPE of Pose2Mesh on Human3.6M by testing only on the frontal camera set, following the previous works~\cite{kanazawa2018hmr,kolotouros2019cmr,kolotouros2019spin}.
In addition, we used 2D human poses estimated from DarkPose~\cite{zhang2020distribution} as input for 3DPW evaluation, which improved HRNet~\cite{sun2019deep}.

\begin{table}[t]
\setlength{\tabcolsep}{1pt}
\caption{The accuracy comparison between state-of-the-art methods and Pose2Mesh on Human3.6M and 3DPW. Different train sets are used. $*$ means synthetic data from AMASS~\cite{mahmood2019amass} is used. Refer to Section~\ref{sec:amass} for more details.}
\centering
\scalebox{1.0}{
\begin{tabular}{C{4.0cm}|C{1.5cm}C{1.7cm}|C{1.5cm}C{1.7cm}C{1.5cm}}
\specialrule{.1em}{.05em}{.05em}
\multirow{ 2}{*}{method} & \multicolumn{2}{c|}{Human3.6M} & \multicolumn{2}{c}{3DPW} \\
                       & MPJPE & PA-MPJPE & MPJPE & PA-MPJPE & MPVPE\\ \hline
SMPLify~\cite{bogo2016keep} & - & 82.3 & - & - & -\\
Lassner~et al.~\cite{lassner2017unite} &- & 93.9 & - & - & -\\
HMR~\cite{kanazawa2018hmr} & 88.0 & 56.8 & - & 81.3 & - \\
NBF~\cite{omran2018neural} & - & 59.9 & - & - & -\\
Pavlakos~et al.~\cite{pavlakos2018l3d} & - & 75.9 & - & - & - \\
Kanazawa~et al.~\cite{kanazawa2019learning} & - & 56.9 & - & 72.6 & - \\
GraphCMR~\cite{kolotouros2019cmr} & - & 50.1 & - & 70.2 & - \\
Arnab~et al.~\cite{arnab2019exploiting} &77.8 & 54.3 & - & 72.2 & - \\
SPIN~\cite{kolotouros2019spin} & - & \textbf{41.1} & - & 59.2 & 116.4 \\
\textbf{Pose2Mesh (Ours)} & \textbf{64.9} & 46.3 & \textbf{88.9} & 58.3 & 106.3 \\ \textbf{Pose2Mesh (Ours)$*$} & - & - & 89.5 & \textbf{56.3} & \textbf{105.3}\\
\specialrule{.1em}{.05em}{.05em}
\end{tabular}
}
\label{table:compare_h36m_3dpw_different_dataset}
\end{table}

\begin{figure*}[t]
\begin{center}
\includegraphics[width=0.8\linewidth]{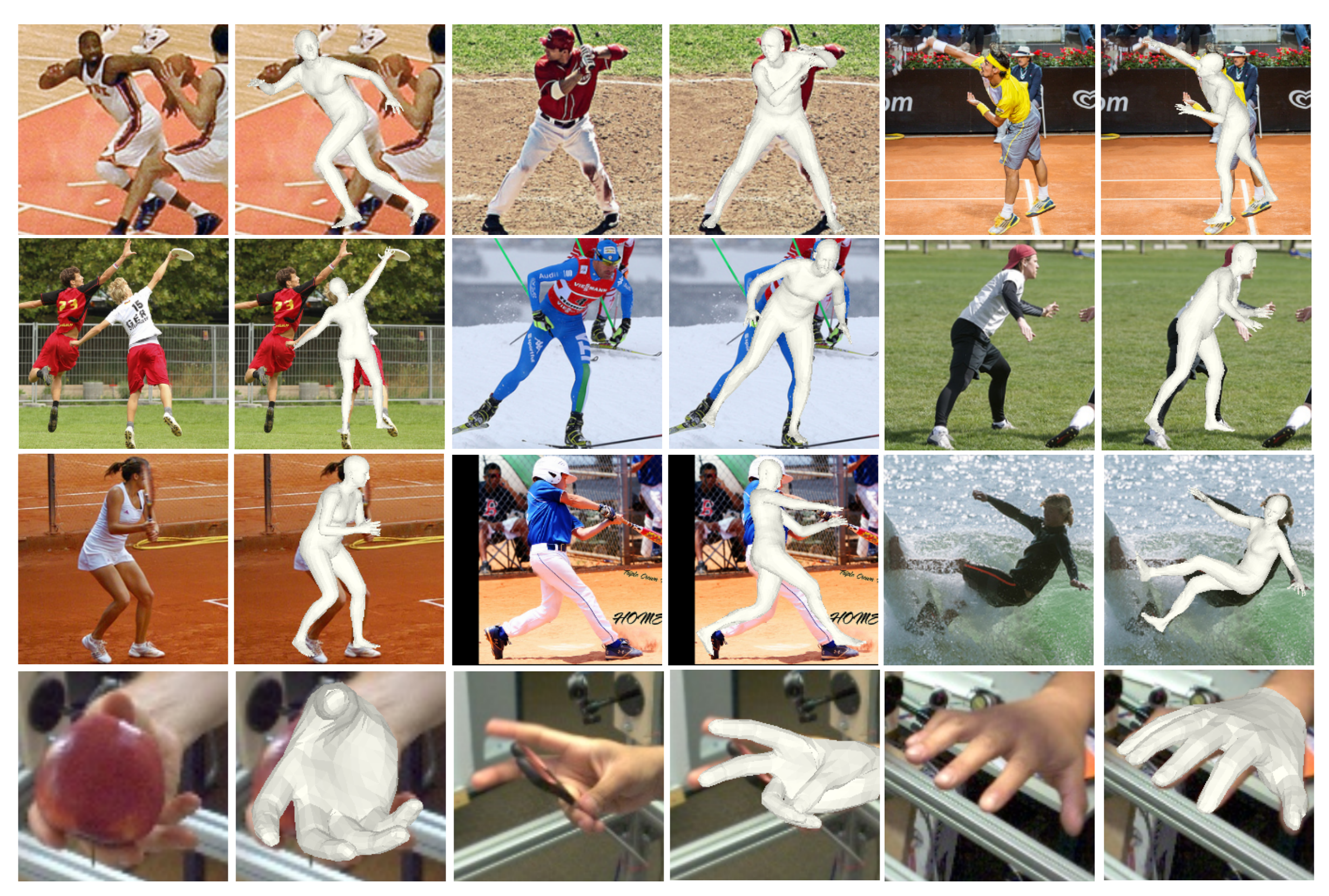}
\end{center}
\vspace{-1em}
   \caption{
   Qualitative results of the proposed Pose2Mesh. First to third rows: COCO, fourth row: FreiHAND.
   }
\label{fig:qualitative_coco_freihand}
\vspace{-1.5em}
\end{figure*}

Figure~\ref{fig:qualitative_coco_freihand} shows the qualitative results on COCO validation set and FreiHAND test set.
Our Pose2Mesh outputs visually decent human meshes without post-processing, such as model fitting~\cite{kolotouros2019cmr}.
More qualitative results can be found in the supplementary material.

\vspace{-2mm}

\vspace{-2mm}
\section{Discussion}
Although the proposed system benefits from the homogeneous geometric property of input 2D poses from different domains, it could be challenging to recover various 3D \emph{shapes} solely from the pose.
While it may be true, we found that the 2D pose still carries necessary information to reason the corresponding 3D shape.
In the literature, SMPLify~\cite{bogo2016keep} has experimentally verified that under the canonical body pose, utilizing 2D pose significantly drops the body shape fitting error compared to using the mean body shape.
We show that Pose2Mesh can recover various body shapes from the 2D pose in the supplementary material.

\vspace{-2mm}
\section{Conclusion}
We propose a novel and general system, Pose2Mesh, for 3D human mesh and pose estimation from a 2D human pose. 
The input 2D pose enables the system to benefit from the 3D data captured from the controlled settings without the appearance domain gap issue.
The model-free approach using GraphCNN allows it to fully exploit mesh topology, while avoiding the representation issues of the 3D rotation parameters.
We plan to enhance the shape recover capability of Pose2Mesh using denser keypoints or part segmentation, while maintaining the above advantages.

\vspace*{+2mm}

\noindent\textbf{Acknowledgements.} This work was supported by IITP grant funded by the Ministry of Science and ICT of Korea (No.2017-0-01780), and Hyundai Motor Group through HMG-SNU AI Consortium fund (No. 5264-20190101).

\clearpage

\begin{center}
\textbf{\large Supplementary Material of \\ \enquote{Pose2Mesh: Graph Convolutional Network for\\3D Human Pose and Mesh Recovery\\from a 2D Human Pose}}
\end{center}

In this supplementary material, we present more experimental results that could not be included in the main manuscript due to the lack of space.

\section{Qualitative results}
\subsection{Shape recovery}
\begin{figure}[!hbt]
\setlength\belowcaptionskip{-5ex}
\centerline{
\includegraphics[width=0.9\linewidth,trim={4pt 4pt 4pt 4pt}]{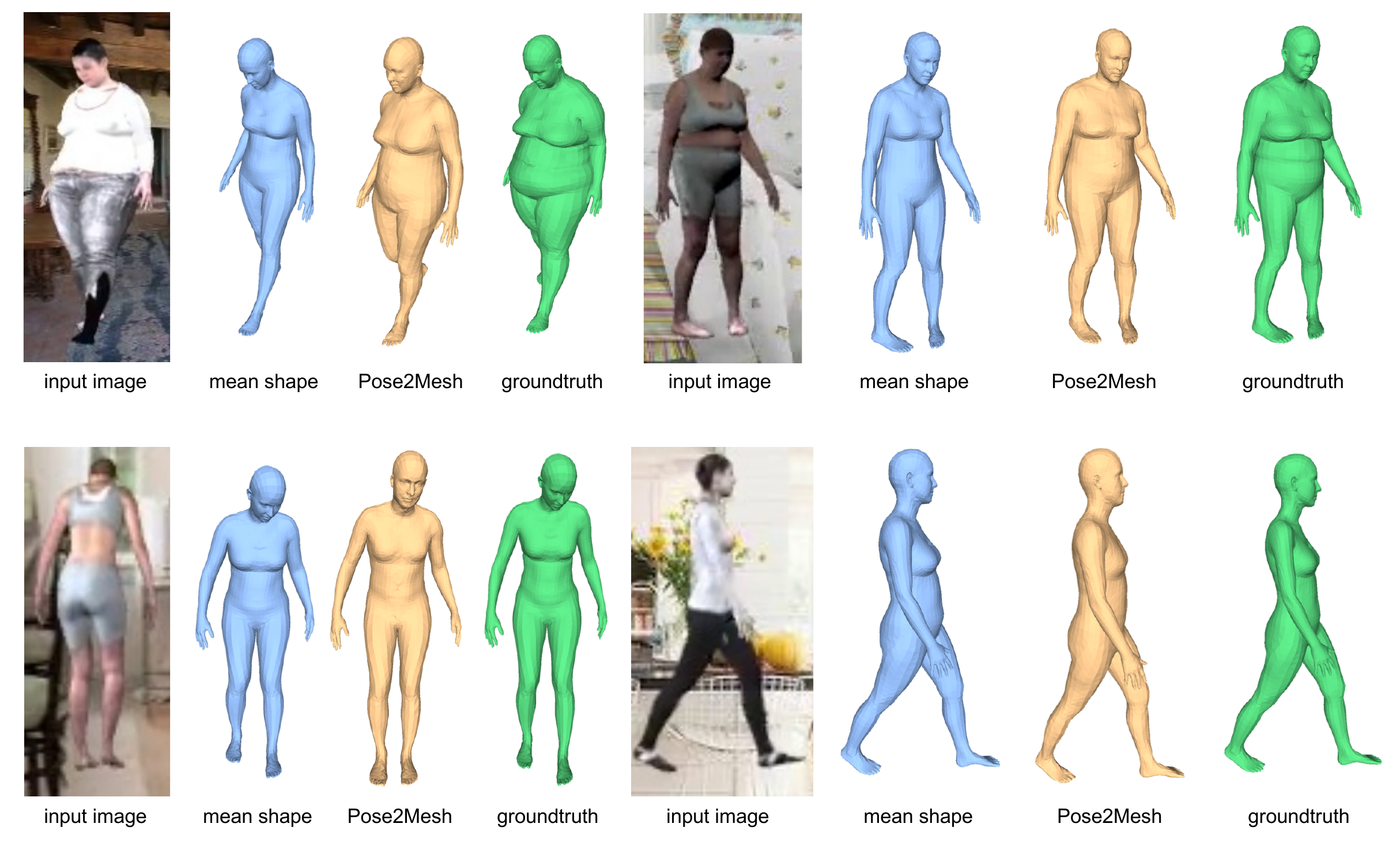}}
\vspace*{-1mm}
\caption
{
The Pose2Mesh predictions compared with the groundtruth mesh, and the mesh decoded from groundtruth pose parameters and the mean shape parameters.
}
\label{fig:surreal}
\end{figure}

We trained and tested Pose2Mesh on SURREAL~\cite{varol2017surreal}, which have various samples in terms of the body shape, to verify the capability of shape recovery.
As shown in Figure~\ref{fig:surreal}, Pose2Mesh can recover a 3D body shape corresponding to an input image, though not perfectly.
The shape features of individuals, such as the bone length ratio and fatness, are expressed in the outputs of Pose2Mesh.
This implies that the information embedded in joint locations (\textit{e.g.} the distance between hip joints) carries a certain amount of shape cue.

\subsection{Additional results}
\begin{figure}[!hbt]
\setlength\belowcaptionskip{-5ex}
\centerline{
\includegraphics[width=0.9\linewidth,trim={4pt 4pt 4pt 4pt}]{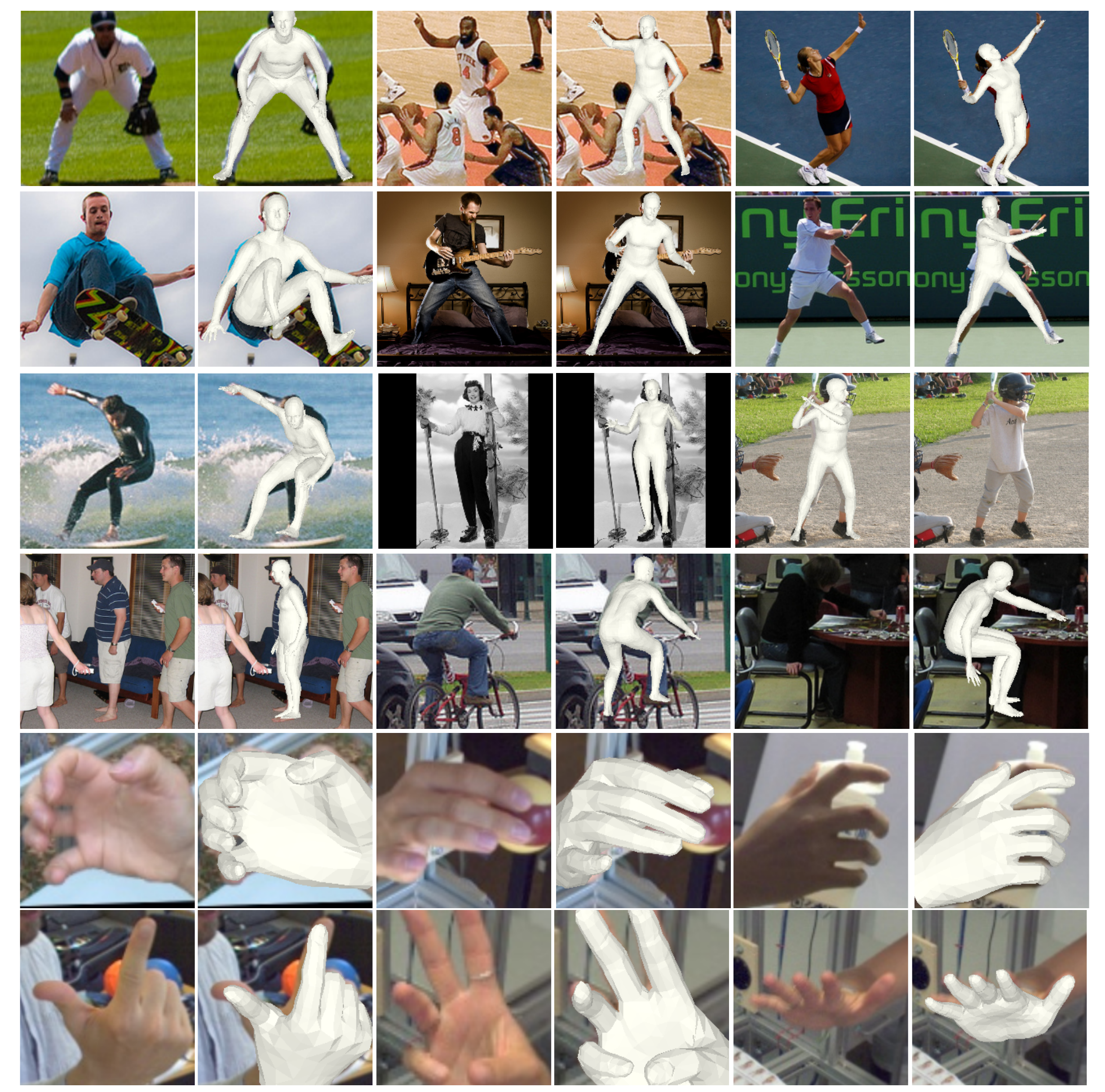}}
\vspace*{-1mm}
\caption
{
Additional qualitative results on COCO and FreiHAND.
}
\label{fig:more_quality_figures.pdf}
\end{figure}

Here, we present more qualitative results on COCO~\cite{lin2014mscoco} validation set and FreiHAND~\cite{chris2019Freihand} test set in Figure~\ref{fig:more_quality_figures.pdf}.
The images at the fourth row show some of the failure cases. 
Although the people on the first and second images appear to be overweight, the predicted meshes seem to be closer to the average shape.
The right arm pose of the mesh in the third column is bent, though it appears straight.

\subsection{Comparison with the state-of-the-art}
\begin{figure}
\begin{center}
\includegraphics[width=1.0\linewidth]{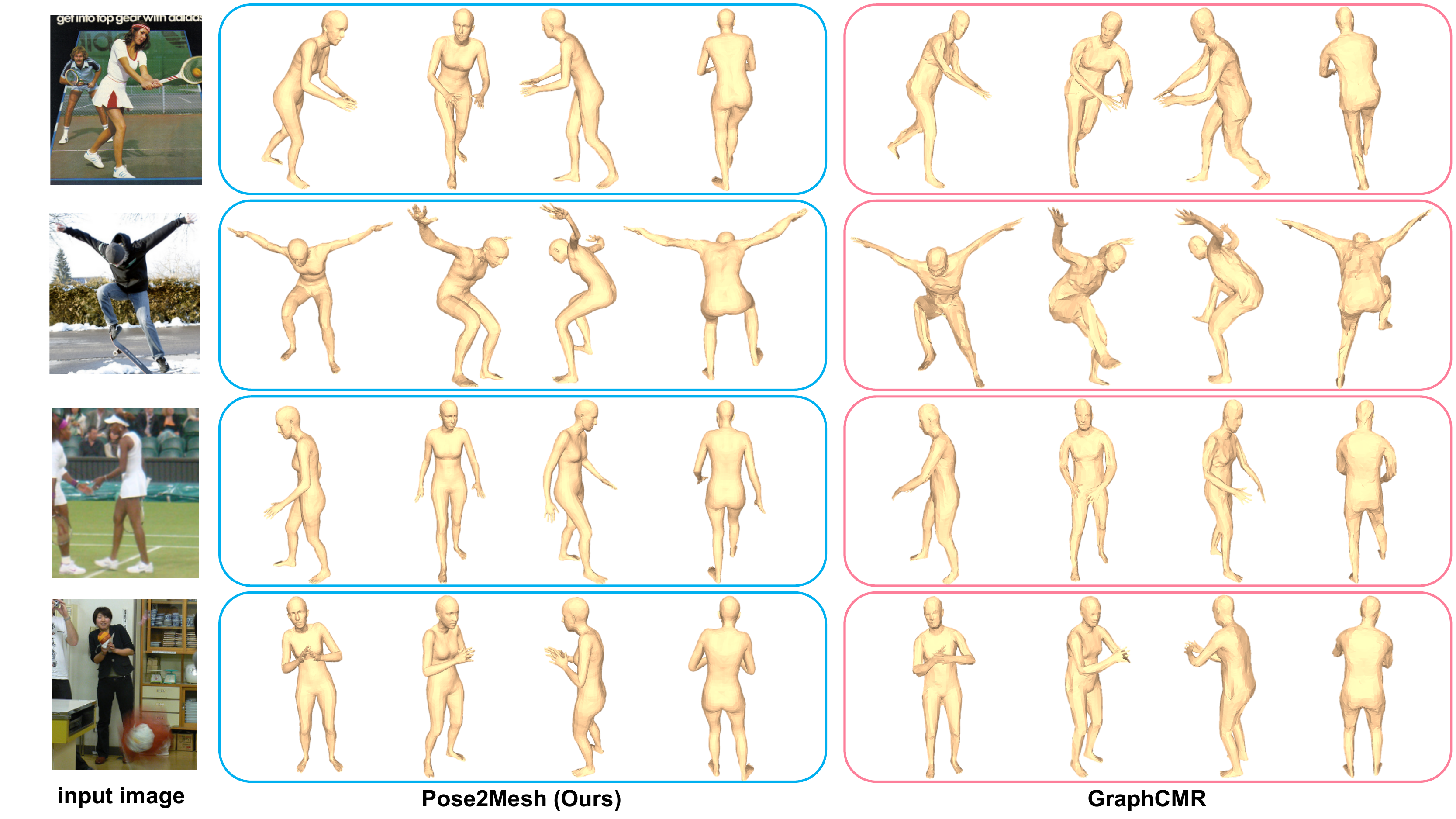}
\end{center}
\vspace*{-3mm}
   \caption{
    The mesh quality comparison between our Pose2Mesh and GraphCMR~\cite{kolotouros2019cmr}.
   }
\label{fig:compare}
\end{figure}

We present the qualitative comparison between our Pose2Mesh and GraphCMR~\cite{kolotouros2019cmr} in Figure~\ref{fig:compare}.
We regard GraphCMR as a suitable comparison target, since it is also the model-free method and regresses coordinates of human mesh defined by SMPL~\cite{loper2015smpl} using GraphCNN like ours.
As the figure shows, our Pose2Mesh provides much more visually pleasant mesh results than GraphCMR.
Based on the loss function analysis in Section 7 and the visual results of GraphCMR, we conjecture that the surface losses such as the normal loss and the edge loss are the reason for the difference. 

\section{Details of PoseNet}
\subsection{Network architecture}
\begin{figure}[!hbt]
\centerline{
\includegraphics[width=0.8\linewidth,trim={4pt 4pt 4pt 4pt}]{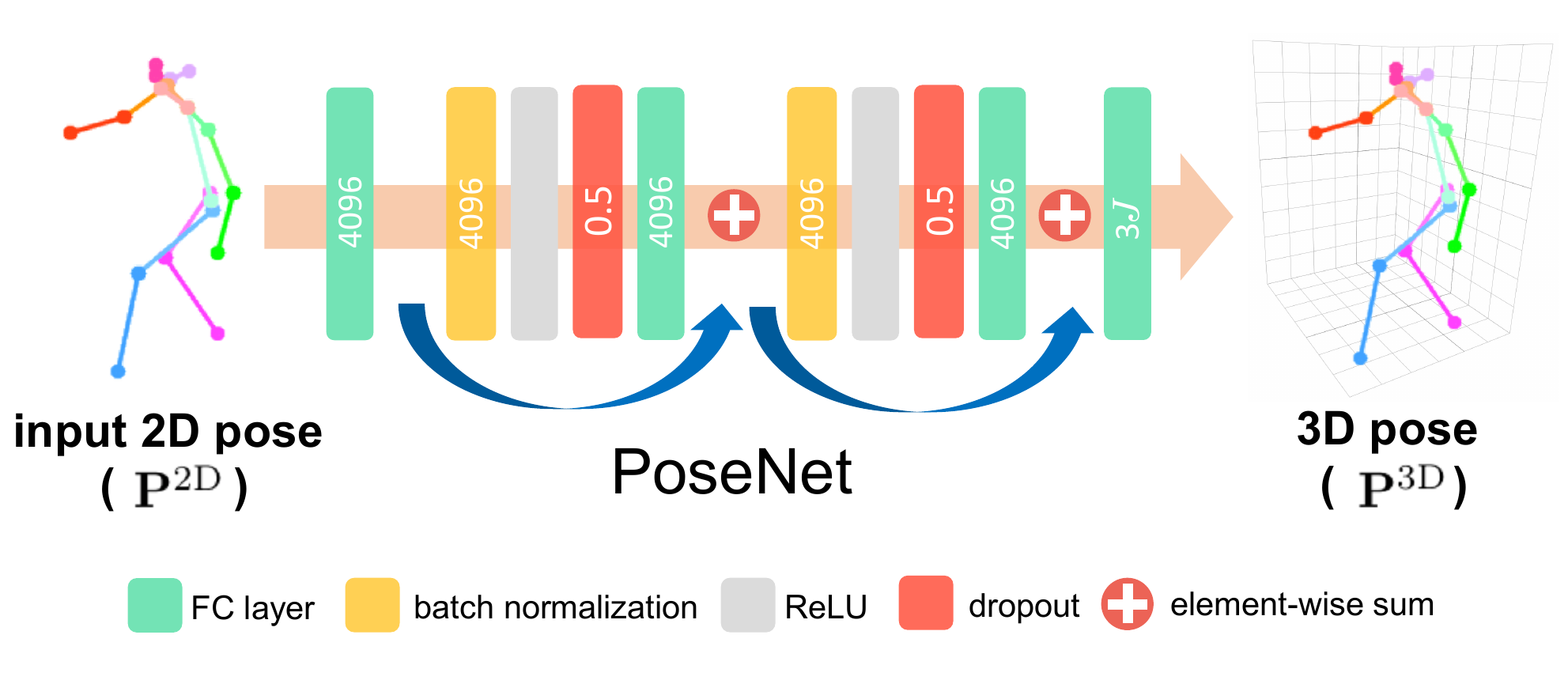}}
\caption
{
The detailed network architecture of PoseNet.
}
\label{fig:posenet}
\end{figure}

Figure~\ref{fig:posenet} shows the detailed network architecture of PoseNet.
First, the normalized input 2D pose vector is converted to a 4096-dimensional feature vector by a fully-connected layer.
Then, it is fed to the two residual blocks, where each block consists of a fully connected layer, 1D batch normalization, ReLU activation, and the dropout.
The dimension of the feature map in the residual block is 4096, and the dropout probability is set to 0.5.
Finally, the output from the residual block is converted to $(3J)$-dimensional vector, the 3D pose vector, by a fully-connected layer.
The 3D pose vector represents the root-relative 3D pose coordinates.

\subsection{Accuracy of PoseNet}
We present MPJPE and PA-MPJPE of PoseNet on the benchmarks in Table~\ref{table:posenet_error}.
For the Human3.6M benchmark~\cite{ionescu2014human3}, 14 common joints out of 17 Human3.6M defined joints are evaluated following~\cite{kanazawa2018hmr,pavlakos2018l3d,kolotouros2019cmr,kolotouros2019spin}.
For the 3DPW benchmark~\cite{von20183dpw}, COCO defined 17 joints are evaluated and $\mathcal{J}\mathbf{M}$ from the groundtruth SMPL meshes are used as groundtruth. 
The 2D pose outputs from~\cite{sun2018integral} and~\cite{sun2019deep} are taken as test inputs on Human3.6M and 3DPW respectively.
For the FreiHAND benchmark, only FreiHAND train set is used during training, and 21 MANO~\cite{romero2017mano} hand joints are evaluated by the official evaluation website.
The 2D pose outputs from~\cite{sun2019deep} are taken as test inputs.

\begin{table}[!hbt]
\centering
\setlength\tabcolsep{1.0pt}
\def\arraystretch{1.0}
\caption{The MPJPE and PA-MPJPE of PoseNet on each benchmark. }
\begin{tabular}{C{2.8cm}|C{1.3cm}C{1.8cm}|C{1.3cm}C{1.8cm}}
\specialrule{.1em}{.05em}{.05em}
train set & \multicolumn{2}{c|}{ Human3.6M} & \multicolumn{2}{c}{ Human3.6M + COCO} \\ \hline
benchmark &  MPJPE &  PA-MPJPE &  MPJPE &  PA-MPJPE \\ \hline
 Human3.6M &  65.1 & 48.4 & 66.7 & 48.9 \\
 3DPW & 105.0 & 62.9 & 99.2 & 61.0 \\
 \specialrule{.1em}{.05em}{.05em}
\end{tabular}
\newline
\vspace*{0.1 cm}
\newline

\begin{tabular}{C{2.8cm}|C{1.8cm}}
\specialrule{.1em}{.05em}{.05em}
benchmark &  PA-MPJPE \\ \hline
 FreiHAND &  8.56 \\
 \specialrule{.1em}{.05em}{.05em}
\end{tabular}

\label{table:posenet_error}
\end{table}

\section{Pre-defined joint sets and graph structures}
\begin{figure}[!hbt]
\setlength\belowcaptionskip{-2ex}
\centerline{
\includegraphics[width=0.8\linewidth,trim={4pt 4pt 4pt 4pt}]{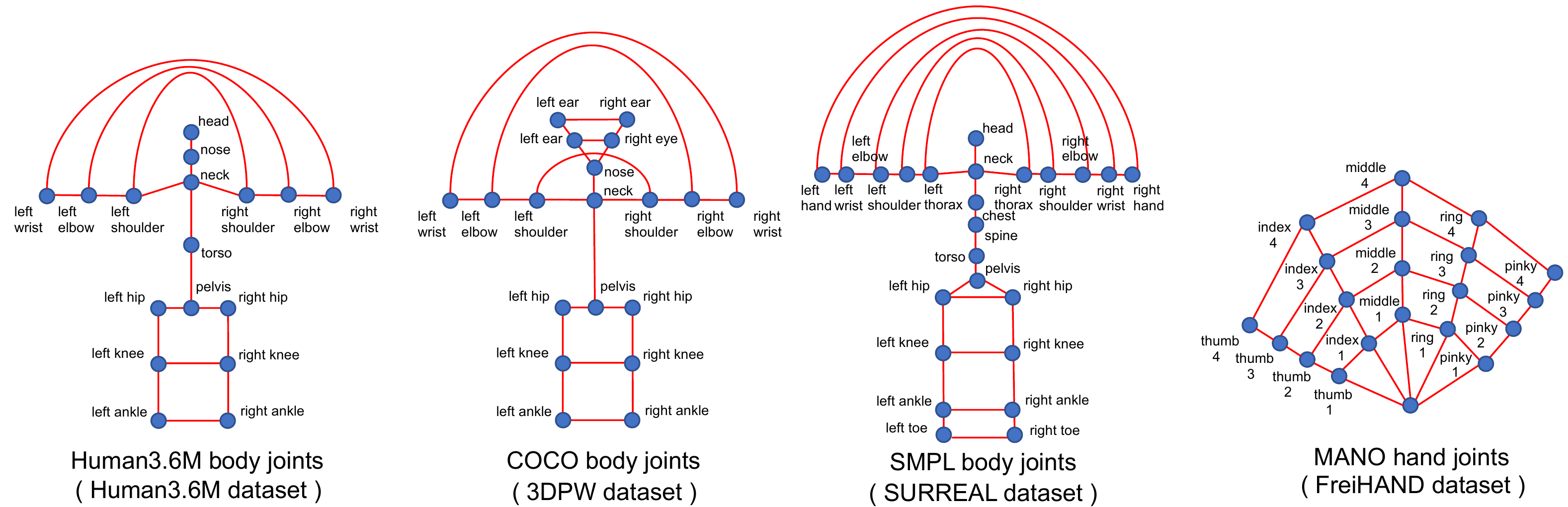}}
\vspace*{-1mm}
\caption
{
The joint sets and graph structures of each dataset that are used in Pose2Mesh.
}
\label{fig:skeleton}
\end{figure}

We use different pre-defined joint sets and graph structures for Human3.6M, 3DPW, SURREAL, and FreiHAND benchmarks, as shown in Figure~\ref{fig:skeleton}.
To be specific, we employ Human3.6M body joints, COCO body joints, SMPL body joints, MANO hand joints for Human3.6M, 3DPW, SURREAL, FreiHAND benchmarks, respectively, in both training and testing stages.
For the COCO joint set, we additionally define pelvis and neck joints that connect the upper body and lower body.
The pelvis and neck coordinates are calculated as the middle point of right-left hips and right-left shoulders, respectively.

\section{Pseudo-groundtruth SMPL parameters of Human3.6M}
Mosh~\cite{loper2014mosh} method can compute SMPL parameters from the marker data in Human3.6M.
Since Human3.6M does not provide 3D mesh annotations, most of the previous 3D pose and mesh estimation papers~\cite{kanazawa2018hmr,pavlakos2018l3d,kolotouros2019cmr,kolotouros2019spin} used the SMPL parameters obtained by Mosh method as the groundtruth for the supervision.
However, due to the license issue, the SMPL parameters are not currently available.
Furthermore, the source code of Mosh is not publicly released.

For the 3D mesh supervision, we alternatively obtain groundtruth SMPL parameters by applying SMPLify-X~\cite{pavlakos2019expressive} on the groundtruth 3D joint coordinates of Human3.6M.
Although the obtained SMPL parameters are not perfectly aligned to the groundtruth 3D joint coordinates, we confirmed that the error of the SMPLify-X is much less than those of current state-of-the-art 3D human pose estimation methods, as shown in Table~\ref{table:smplify-x_error}.
Thus, we believe using SMPL parameters obtained by SMPLify-X as groundtruth is reasonable.
For the fair comparison, all the previous works and our system are trained on our SMPL parameters from SMPLify-X.

During the fitting process of SMPLify-X, we adopted a neutral gender SMPL body model.
However, we empirically found that the fitting process produces gender-specific body shapes, which correspond to each subject.
As a result, since most of the subjects in the training set of Human3.6M are female, our Pose2Mesh trained on Human3.6M tends to produce female body shape meshes.
We tried to fix the identity code of the SMPL body model obtained from the T-pose; however, it produces higher errors.
Thus, we did not fix the identity code for each subject.

\begin{table}
\centering
\setlength\tabcolsep{1.0pt}
\def\arraystretch{1.1}
\caption{The MPJPE comparison between SMPLify-X fitting results and state-of-the-art 3D human pose estimation methods. \enquote{*} takes multi-view RGB images as inputs.}
\scalebox{1.0}{
\begin{tabular}{C{4.5cm}|C{2.0cm}C{2.0cm}}
\specialrule{.1em}{.05em}{.05em}
methods & MPJPE  \\ \hline
Moon~et al.~\cite{moon2019camera} & 53.3  \\ 
Sun~et al.~\cite{sun2018integral} & 49.6  \\
Iskakov~et al.~\cite{iskakov2019learnable}*  & 20.8 \\
SMPLify-X from GT 3D pose & \textbf{13.1}  \\ 
 \specialrule{.1em}{.05em}{.05em}
\end{tabular}
}
\vspace*{-7mm}
\label{table:smplify-x_error}
\end{table}

\section{Synthetic data from AMASS}\label{sec:amass}
We leverage additional synthetic data from AMASS~\cite{mahmood2019amass} to boost the performance of Pose2Mesh.
AMASS is a new database that unifies 15 different optical marker-based mocap datasets within a common framework.
It created SMPL parameters from mocap data by a method named Mosh++. 
We used CMU~\cite{cmumocap} and BML-Movi~\cite{ghorbani2020movi} from the database in training PoseNet and only CMU in training Pose2Mesh.

To be specific, we generated paired 2D pose-3D mesh data by projecting a 3D pose obtained from a mesh to the image plane, using camera parameters from Human3.6M.
As shown in Table~\ref{table:accumulate_data}, when AMASS is added, both the joint error and surface error decrease.
Exploiting AMASS data in this fashion is not possible for~\cite{kanazawa2018hmr},~\cite{kolotouros2019cmr}, and~\cite{kolotouros2019spin}, since they need pairs of image and 2D/3D annotations.

\begin{table}
\setlength{\tabcolsep}{1pt}
\def\arraystretch{1.1}
\centering
\resizebox{1.0\linewidth}{!}{\begin{minipage}{\linewidth}
\centering
\setlength{\belowcaptionskip}{3pt plus 3pt minus 2pt}

\centering
\setlength{\belowcaptionskip}{3pt plus 3pt minus 2pt}
\caption{The MPJPE and MPVPE of our Pose2Mesh on 3DPW with accumulative training datasets. We used psuedo-groundtruth SMPL parameters of COCO obtained by NeuralAnnot~\cite{moon2020neuralannot}. The 2D pose outputs from~\cite{sun2019deep} are used for input to Pose2Mesh.}
\scalebox{1.0}{
\begin{tabular}{C{5.0cm}|C{1.5cm}C{1.7cm}C{1.5cm}}
\specialrule{.1em}{.05em}{.05em}
 train sets &  MPJPE & PA-MPJPE & MPVPE \\ \hline
 Human3.6M+COCO &  92.6 & 57.9 & 109.4 \\
 Human3.6M+COCO+AMASS &  \textbf{89.5} & \textbf{56.3} & \textbf{105.3} \\
 \specialrule{.1em}{.05em}{.05em}
\end{tabular}
}
\label{table:accumulate_data}
\end{minipage}}
\end{table}

\section{Synthesizing the input 2D poses in the training stage}
\subsection{Detailed description of the synthesis}
As described in Section 4.1 of the main manuscript,
we synthesize the input 2D poses by adding randomly generated errors on the groundtruth 2D poses in the training stage.
For this, we generate errors following Chang~et al.~\cite{chang2020abs} and Moon~et al.~\cite{Moon_2019_CVPR_PoseFix} for Human3.6M and COCO body joint sets, respectively.
On the other hand, for FreiHAND benchmark, we used detection outputs from~\cite{sun2019deep} on the training set as the input poses in the training stage, since there are no verified synthetic errors for the hand joints.

\subsection{Effect of synthesizing the input 2D poses}
To demonstrate the validity of the synthesizing process, we compare MPJPE and PA-MPJPE of Pose2Mesh trained with the groundtruth 2D poses, and the synthesized input 2D poses in Table~\ref{table:syn_error}.
For Human3.6M, only Human3.6M train set is used for the training, and for 3DPW benchmark, Human3.6M and COCO are used for the training.
The test 2D input poses used in Human3.6M and 3DPW evaluation are outputs from Integral Regression~\cite{sun2018integral} and HRNet~\cite{sun2019deep} respectively, using groundtruth bounding boxes.
Apparently, when our Pose2Mesh is trained with the synthesized input 2D poses, Pose2Mesh performs far better on both benchmarks.
This proves that the synthesizing process makes Pose2Mesh more robust to the errors in the input 2D poses and increases the estimation accuracy.

\begin{table}[!hbt]
\setlength{\tabcolsep}{1pt}
\def\arraystretch{1.1}
\centering
\caption{The MPJPE and PA-MPJPE comparison according to input type in the training stage.}
\scalebox{0.9}{
\begin{tabular}{C{4.5cm}|C{1.5cm}C{2.1cm}|C{1.5cm}C{2.1cm}}
\specialrule{.1em}{.05em}{.05em}
\multirow{ 2}{*}{input pose when training} & \multicolumn{2}{c|}{ Human3.6M} & \multicolumn{2}{c}{ 3DPW} \\
                       &  MPJPE &  PA-MPJPE &  MPJPE &  PA-MPJPE \\ \hline
 2D pose GT &  70.4 &  50.6 &  153.7 &  94.4 \\
 \textbf{2D pose synthesized (Ours)} &  \textbf{64.9} & \textbf{48.7} &  \textbf{91.4} &  \textbf{60.1} \\
 \specialrule{.1em}{.05em}{.05em}
\end{tabular}
}
\label{table:syn_error}
\end{table}

\section{Train/test with groundtruth input poses}
We present the upper bounds of Pose2Mesh, PoseNet, and MeshNet on Human3.6M and 3DPW benchmarks by training and testing with groundtruth input poses in Table~\ref{table:upper_bound}.
Pose2Mesh and PoseNet take the groundtruth 2D pose as an input, while MeshNet takes the groundtruth 3D pose as an input.
As the table shows, the upper bound of Pose2Mesh is similar to that of PoseNet, which implies that the 3D pose errors of Pose2Mesh follow those of PoseNet as analyzed in Section 7.2 of the main manuscript.
In addition, the upper bound of MeshNet indicates that we can recover highly accurate 3D human meshes if we can estimate nearly perfect 3D poses. 

The MPJPE and PA-MPJPE of Pose2Mesh and MeshNet are measured on the 3D pose regressed from the mesh output, while the accuracy of PoseNet is measured on the lifted 3D pose.
For the Human3.6M benchmark, only Human3.6M train set is used to train the network.
For the 3DPW benchmark, Human3.6M, COCO, AMASS train sets are used to train the network. 

\begin{table}[!hbt]
\setlength{\tabcolsep}{1pt}
\def\arraystretch{1.1}
\centering
\caption{The upper bounds of Pose2Mesh, PoseNet, and MeshNet on Human3.6m and 3DPW benchmarks.}
\scalebox{0.9}{
\begin{tabular}{C{4.5cm}|C{1.5cm}C{2.1cm}|C{1.5cm}C{2.1cm}}
\specialrule{.1em}{.05em}{.05em}
\multirow{ 2}{*}{networks} & \multicolumn{2}{c|}{ Human3.6M} & \multicolumn{2}{c}{ 3DPW} \\
                       &  MPJPE &  PA-MPJPE &  MPJPE &  PA-MPJPE \\ \hline
 Pose2Mesh with 2D pose GT &  51.1 &  35.3 &  65.1 &  34.6 \\ 
 PoseNet with 2D pose GT & 50.6 & 41.3 & 66.1 & 43.8 \\
 MeshNet with 3D pose GT &  13.9 &  9.9 &  10.8 &  8.1 \\
 \specialrule{.1em}{.05em}{.05em}
\end{tabular}
}
\label{table:upper_bound}
\end{table}

\section{Effect of each loss function}
\begin{figure}
\begin{center}
\includegraphics[width=1.0\linewidth]{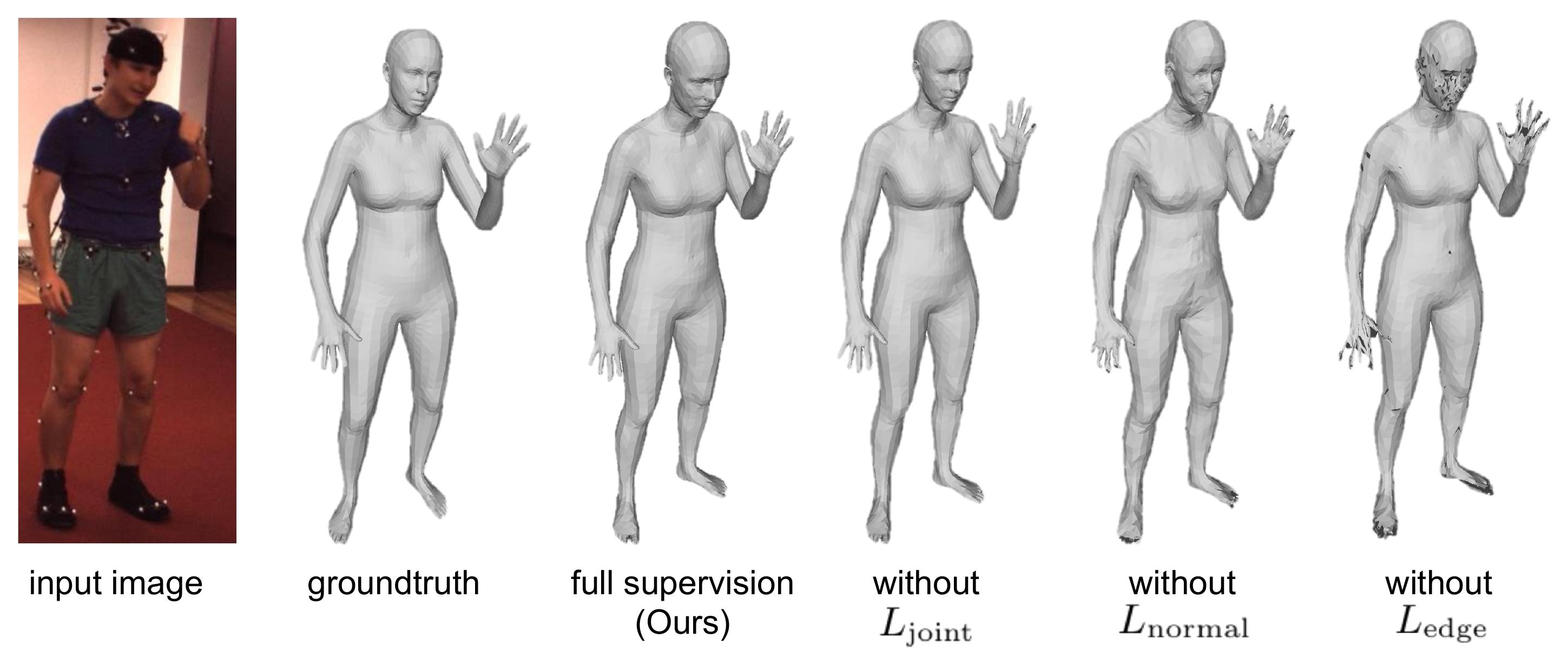}
\end{center}
   \caption{
    Qualitative results for the ablation study on the effectiveness of each loss function.
   }
\label{fig:loss_ablation}
\end{figure}

We analyze the effect of joint coordinate loss $L_\text{joint}$, surface normal loss $L_\text{normal}$, and surface edge loss $L_\text{edge}$ on reconstructing a 3D human mesh in Table~\ref{table:loss_ablation} and Figure~\ref{fig:loss_ablation}.
Human3.6M is used for the training and testing.
As the table shows, training without $L_\text{joint}$ has a relatively distinctive effect on MPJPE and PA-MPJPE, while other settings show numerically negligible differences.
On the other hand, as the figure shows, training without $L_\text{normal}$ or $L_\text{edge}$ clearly decreases the visual quality of the mesh output, while training without $L_\text{joint}$ has nearly no effect on the visual quality of the meshes.
To be specific, training without $L_\text{normal}$ impairs the overall smoothness of the mesh and local details of mouth, hands, and feet.
Similarly, training without $L_\text{edge}$ ruins the details of body parts that have dense vertices, especially mouth, hands, and feet, by making serious artifacts caused by flying vertices.

\begin{table}
\centering
\setlength\tabcolsep{1.0pt}
\def\arraystretch{1.1}
\caption{The MPJPE and PA MPJPE comparison between the networks trained from various combinations of loss functions.}
\scalebox{1.0}{
\begin{tabular}{C{4.0cm}|C{2.0cm}C{2.0cm}}
\specialrule{.1em}{.05em}{.05em}
settings & MPJPE & PA-MPJPE \\ \hline
\textbf{full supervision (Ours)} & 64.9 & 48.7 \\
without $L_\text{joint}$ & 66.9 & 50.1  \\
without $L_\text{normal}$ & \textbf{64.6} & \textbf{48.5}  \\
without $L_\text{edge}$ & 64.8 & 48.7  \\
 \specialrule{.1em}{.05em}{.05em}
\end{tabular}
}
\label{table:loss_ablation}
\end{table}

\clearpage

\bibliographystyle{splncs04}
\bibliography{main}
\end{document}